\documentclass[10pt,twocolumn,letterpaper]{article}

\usepackage{iccv}
\usepackage{times}
\usepackage{epsfig}
\usepackage{graphicx}
\usepackage{amsmath}
\usepackage{amssymb}

\usepackage{subfigure}
\usepackage{pifont}
\usepackage{color, colortbl}

\usepackage{csquotes}

\usepackage{multirow}
\usepackage{graphicx}
\usepackage{booktabs}

\newcommand*{\twoelementtable}[3][l]%
{%
    \begin{tabular}[t]{@{}#1@{}}%
        #2\tabularnewline
        #3%
    \end{tabular}%
}
\usepackage{array}
\usepackage{tabularx}

\usepackage{algorithm}
\usepackage{algorithmic}

\usepackage[pagebackref=true,breaklinks=true,letterpaper=true,colorlinks,bookmarks=false]{hyperref}

\iccvfinalcopy 

\def\iccvPaperID{2530} 

\ificcvfinal\fi

\begin{document}

\title{RDA: Robust Domain Adaptation via Fourier Adversarial Attacking
}

\author{Jiaxing Huang, Dayan Guan, Aoran Xiao, Shijian Lu\thanks{Corresponding author.} \\ 
Singtel Cognitive and Artificial Intelligence Lab for Enterprises,\\
School of Computer Science and Engineering, Nanyang Technological University\\
{\tt\small \{Jiaxing.Huang, Dayan.Guan, Aoran.Xiao, Shijian.Lu\}@ntu.edu.sg}
}

\maketitle
\ificcvfinal\thispagestyle{empty}\fi


\begin{abstract}
Unsupervised domain adaptation (UDA) involves a supervised loss in a labeled source domain and an unsupervised loss in an unlabeled target domain, which often faces more severe overfitting (than classical supervised learning) as the supervised source loss has clear domain gap and the unsupervised target loss is often noisy due to the lack of annotations. This paper presents RDA, a robust domain adaptation technique that introduces adversarial attacking to mitigate overfitting in UDA. We achieve robust domain adaptation by a novel Fourier adversarial attacking (FAA) method that allows large magnitude of perturbation noises but has minimal modification of image semantics, the former is critical to the effectiveness of its generated adversarial samples due to the existence of `domain gaps’. Specifically, FAA decomposes images into multiple frequency components (FCs) and generates adversarial samples by just perturbating certain FCs that capture little semantic information. With FAA-generated samples, the training can continue the `random walk’ and drift into an area with a flat loss landscape, leading to more robust domain adaptation. Extensive experiments over multiple domain adaptation tasks show that RDA can work with different computer vision tasks with superior performance.
\end{abstract}

\section{Introduction}

\begin{figure}[ht]
\centering
\begin{tabular}{p{3.65cm}<{\centering} p{3.65cm}<{\centering}}
w/o FAA & w/ FAA
\\
\end{tabular}
\centering
\begin{minipage}[h]{0.49\linewidth}
\centering\includegraphics[width=.99\linewidth]{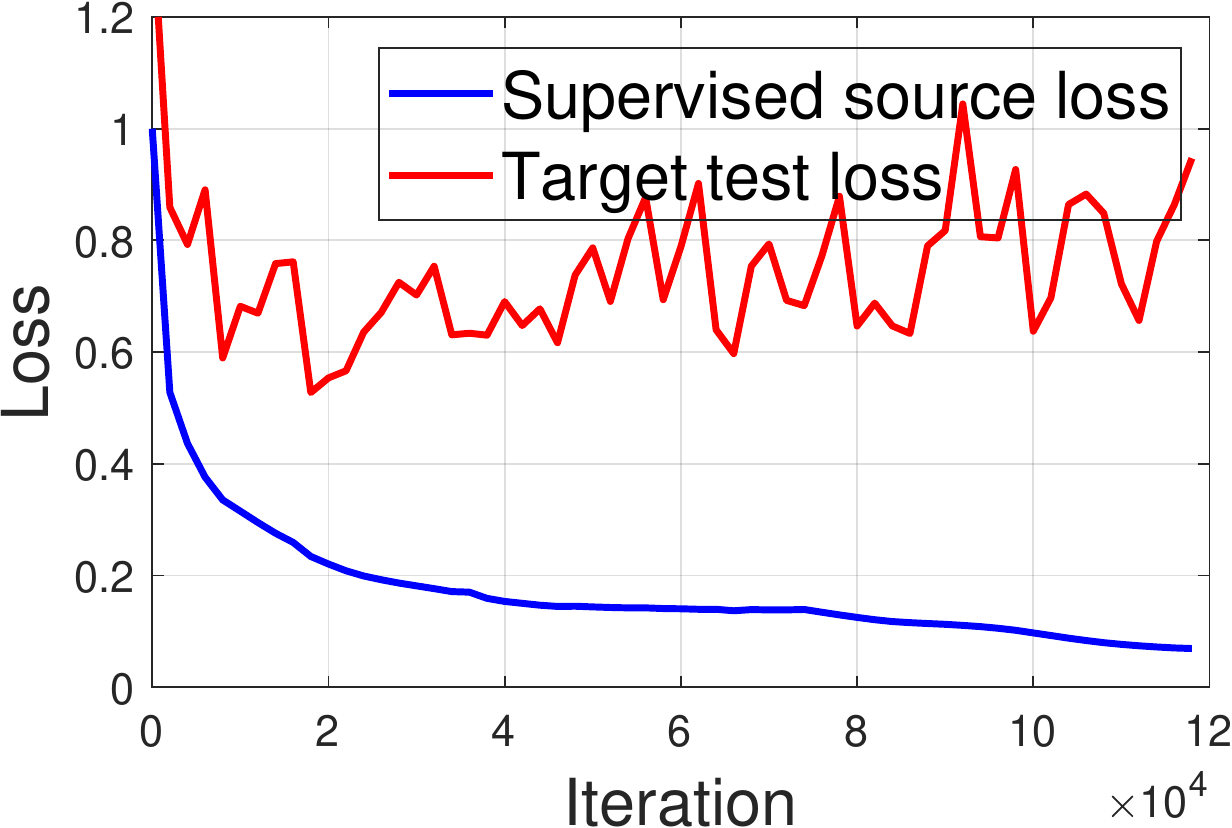}
\end{minipage}
\begin{minipage}[h]{0.49\linewidth}
\centering\includegraphics[width=.99\linewidth]{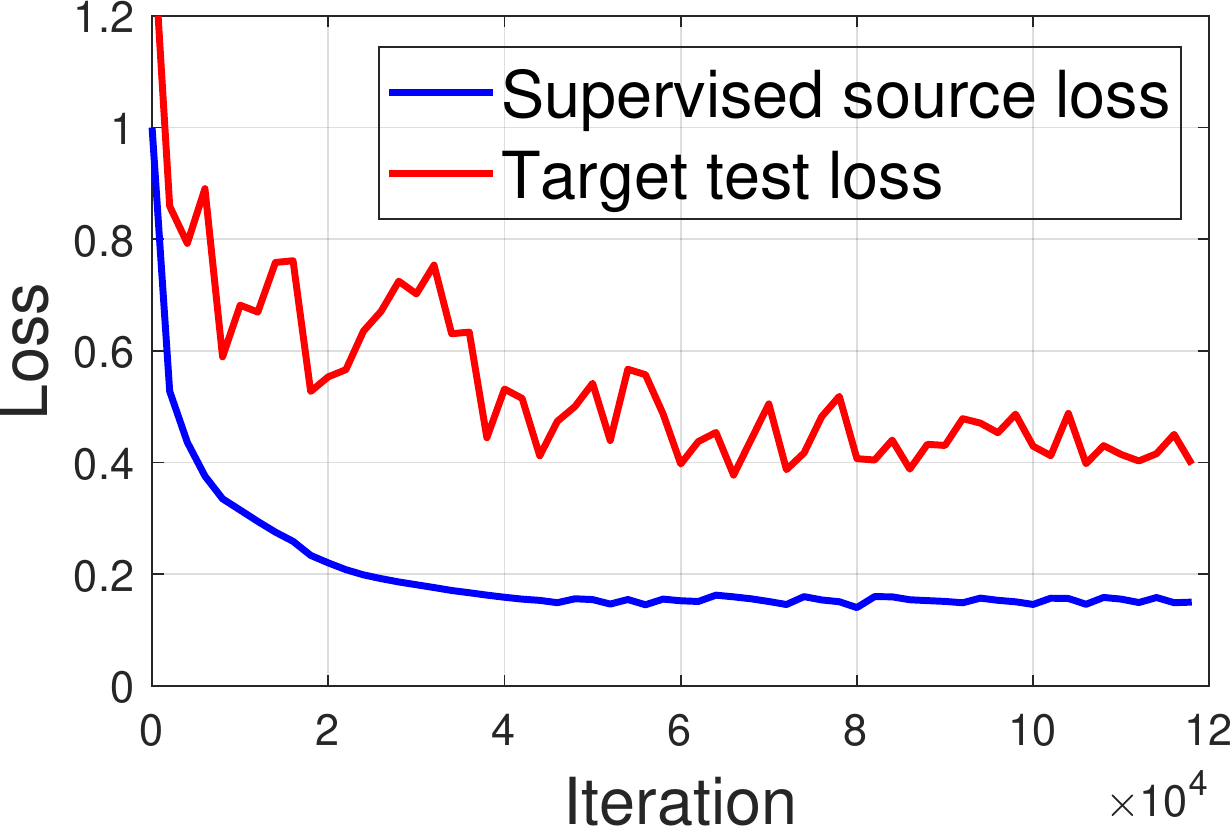}
\end{minipage}
\begin{minipage}[h]{0.49\linewidth}
\centering\includegraphics[width=.99\linewidth]{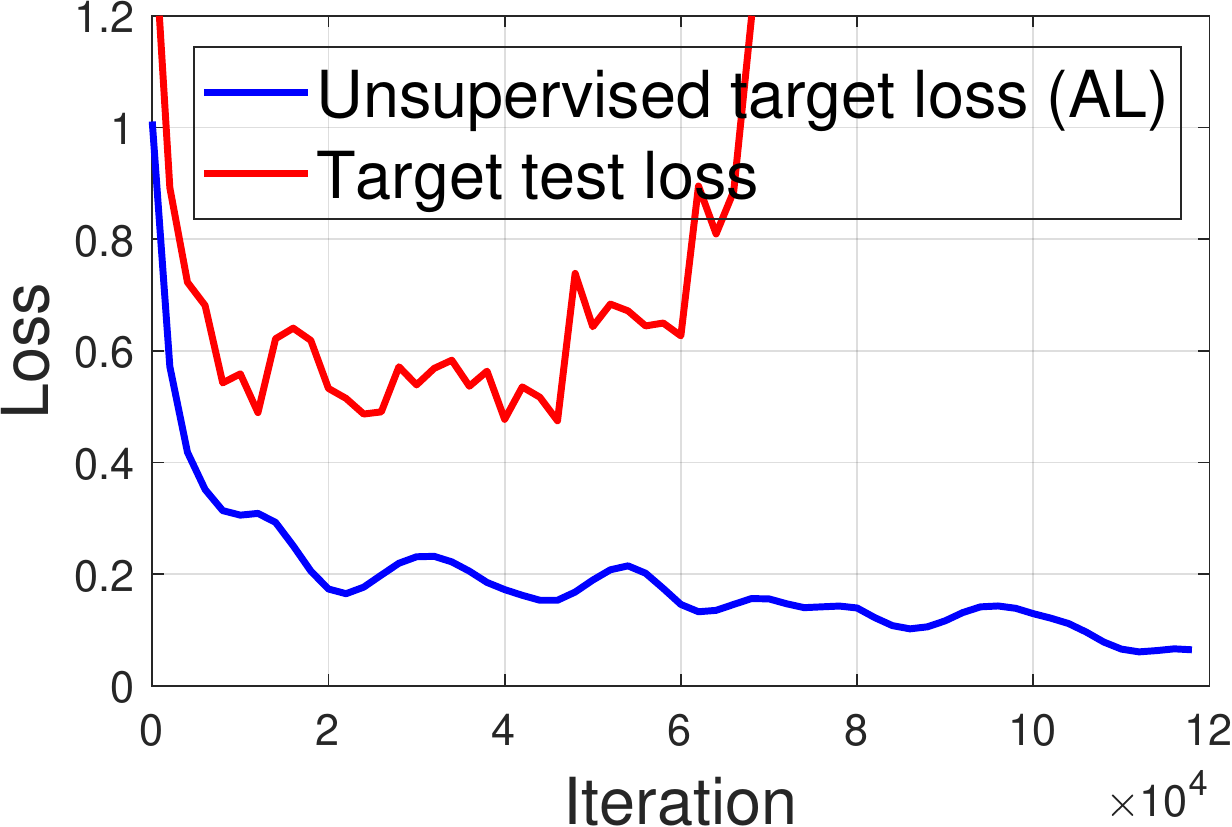}
\end{minipage}
\begin{minipage}[h]{0.49\linewidth}
\centering\includegraphics[width=.99\linewidth]{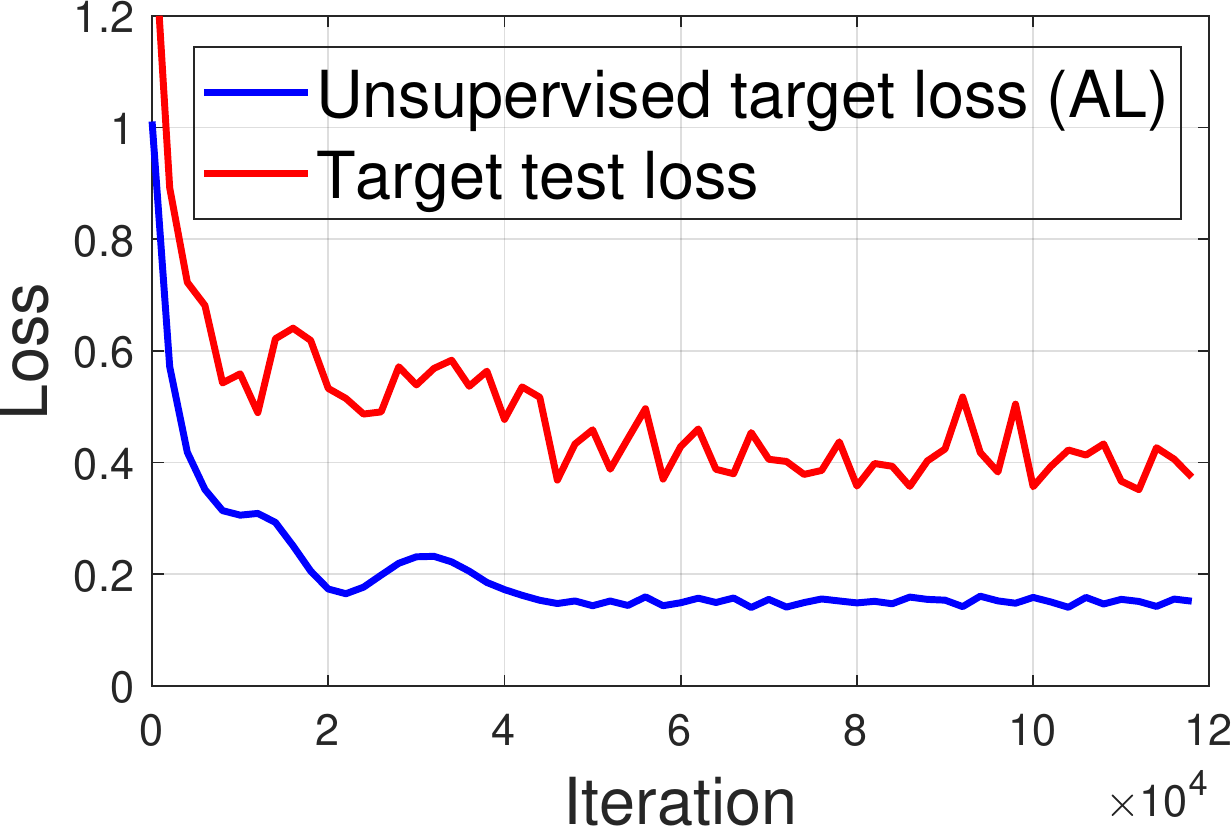}
\end{minipage}
\begin{minipage}[h]{0.49\linewidth}
\centering\includegraphics[width=.99\linewidth]{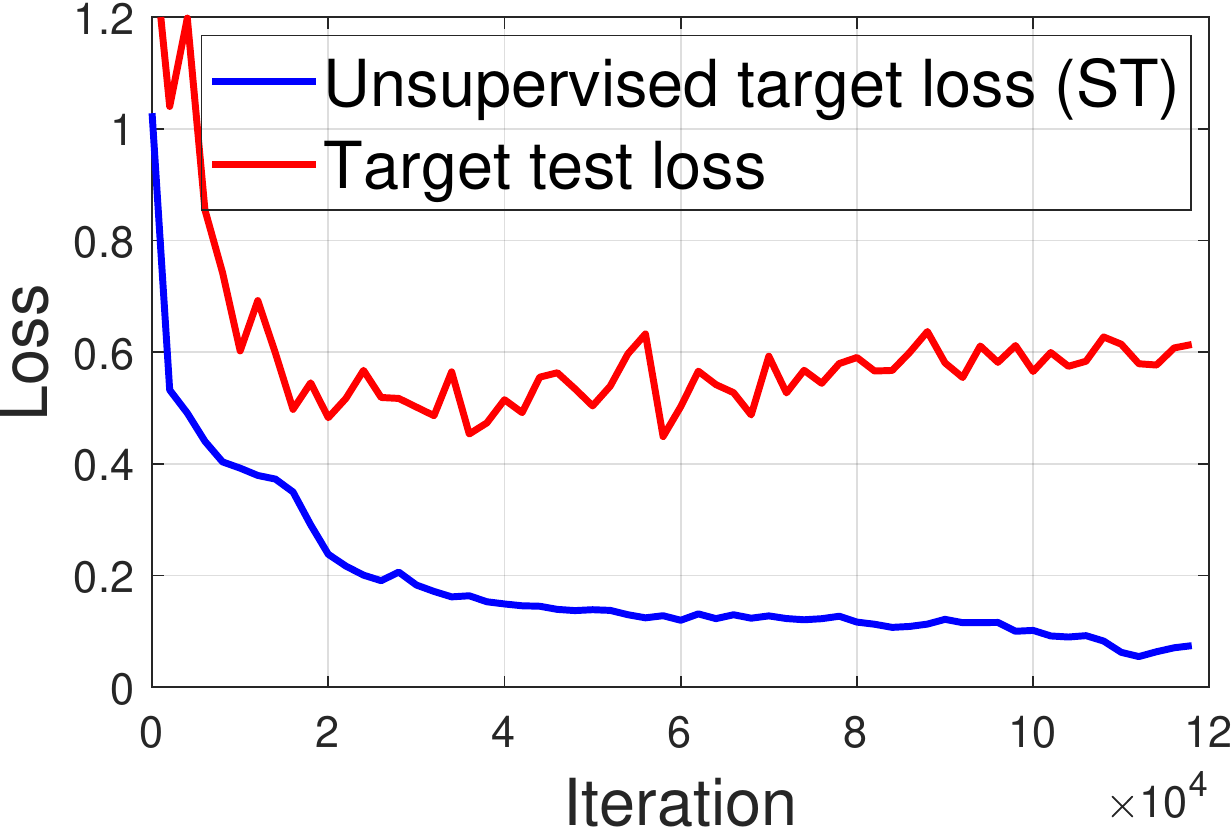}
\end{minipage}
\begin{minipage}[h]{0.49\linewidth}
\centering\includegraphics[width=.99\linewidth]{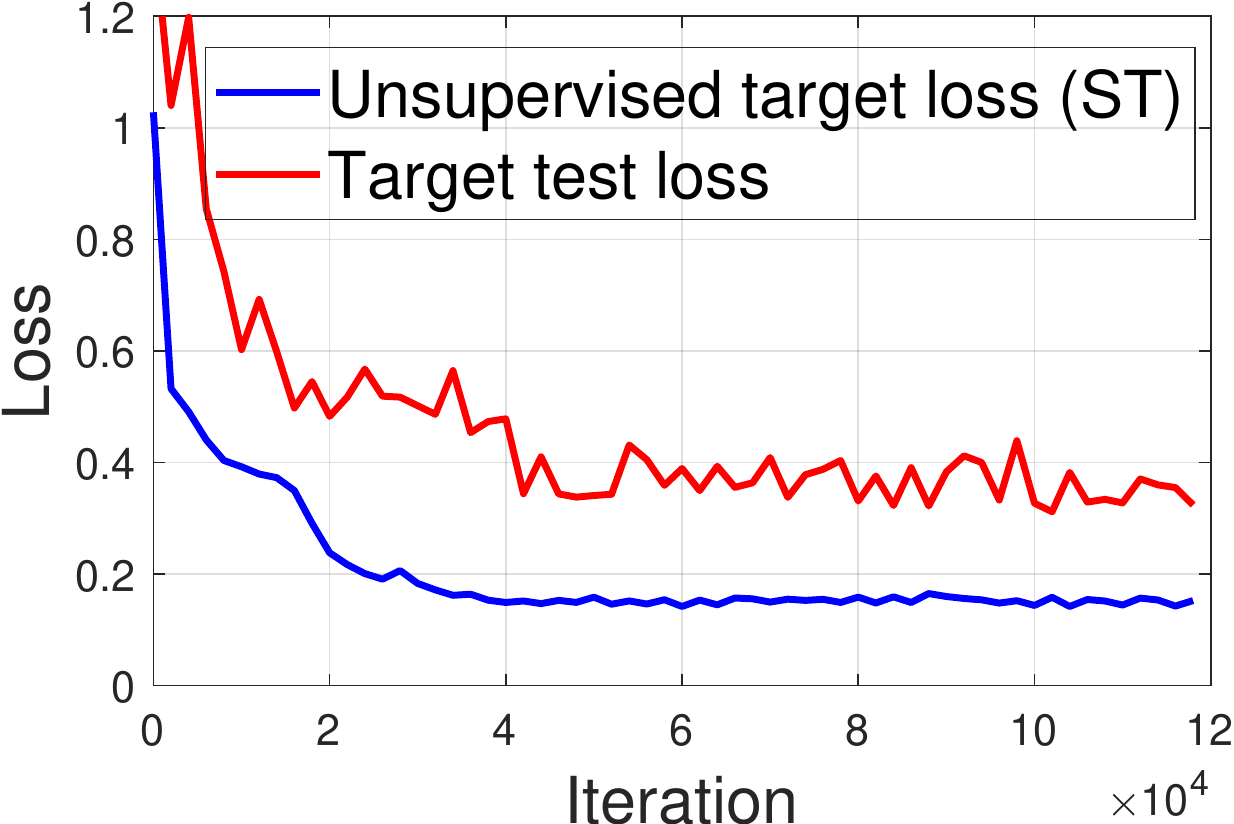}
\end{minipage}
\caption{
Our robust domain adaptation alleviates overfitting effectively: Both supervised learning with source data (row 1) and unsupervised learning with target data (in rows 2 and 3 for adversarial learning and self-training) in unsupervised domain adaptation suffer from clear overfitting as illustrated by decreased training losses (blue curves) vs increased target test losses (red curves) in column 1. Our Fourier adversarial attacking (FAA) generates novel adversarial samples, which regulates the minimization of training losses and alleviates overfitting effectively with decreased target test loss as shown in column 2 (Best viewed in color).}
\label{fig:intro}
\end{figure}

Deep convolutional neural networks (CNNs)~\cite{krizhevsky2012imagenet,simonyan2014very,he2016deep} have defined new state of the arts in various computer vision tasks~\cite{chen2017deeplab,long2015fully,ren2015faster,he2017mask,krizhevsky2012imagenet,simonyan2014very,he2016deep}, but their trained models often over-fit to the training data and experience clear performance drops for data from different sources due to the existence of \textit{domain gaps}. Unsupervised domain adaptation (UDA) has been investigated to address the \textit{domain gaps} by leveraging unlabeled target data. To this end, most existing UDA works~\cite{kang2019contrastive,kang2018deep,tsai2018learning,tzeng2017adversarial, luo2019taking, vu2019advent, Chen_2018_CVPR,chen2018domain} involve supervised losses on source data and unsupervised losses on target data for learning a model that performs well in target domains. However, as illustrated in Fig.~\ref{fig:intro}, these methods often face more severe overfitting (as compared with the classical supervised learning) as supervised source losses in UDA has an extra \textit{domain gap} (for test data in target domains) and unsupervised target losses are often noisy due to the lack of annotations.

Overfitting exists in almost all deep network training, which is undesired and often degrades the generalization of the trained deep network models while applied to new data. One way of identifying whether overfitting is happening is to check whether the generalization gap, \ie, the difference between the test loss and the training loss, is increasing or not~\cite{goodfellow2016deep} as shown in Fig.~\ref{fig:intro}. Various strategies have been investigated to alleviate overfitting through weight regularization~\cite{hanson1988comparing}, dropout~\cite{srivastava2014dropout}, mixup~\cite{zhang2017mixup}, label smoothing~\cite{szegedy2016rethinking}, batch normalization~\cite{ioffe2015batch}, etc. However, all these strategies were designed for supervised and semi-supervised learning where training data and test data usually have very similar distributions. For domain adaptive learning, they do not fit in well due to the negligence of domain gaps that widely exist between data of different domains.

We design a robust domain adaptation technique that introduces a novel Fourier adversarial attacking (FAA) technique to mitigate the overfitting in unsupervised domain adaptation. FAA mitigates overfitting by generating adversarial samples that prevent over-minimization of supervised and unsupervised UDA losses as illustrated in Fig.~\ref{fig:intro}. Specifically, FAA decomposes training images into multiple frequency components (FCs) and only perturbs FCs that capture little semantic information. Unlike traditional attacking that restricts the magnitude of perturbation noises to keep image semantics intact, FAA allows large magnitude of perturbation in its generated adversarial samples but has minimal modification of image semantics. This feature is critical to unsupervised domain adaptation which usually involves clear domain gaps and so requires adversarial sample with large perturbations. By introducing the FAA-generated adversarial samples in training, networks can continue the ``random walk" and avoid over-fitting and drift into an area with a flat loss landscape~\cite{chaudhari2019entropy,keskar2016large,li2017visualizing}, leading to more robust domain adaptation.

The contributions of this work can be summarized in three aspects. \textit{First}, we identify the overfitting issue in unsupervised domain adaptation and introduce adversarial attacking to mitigate overfitting by preventing training objectives from over-minimization. \textit{Second}, we design an innovative Fourier Adversarial Attacking (FAA) technique to generate novel adversarial samples to mitigate overfitting in domain adaptation. FAA is generic which can work for both supervised source loss and unsupervised target losses effectively. \textit{Third}, we conducted extensive experiments over multiple computer vision tasks in semantic segmentation, object detection and image classification. All experiments show that our method mitigates overfitting and improve domain adaption consistently. 

\section{Related Works}

\textbf{Domain Adaptation:} Domain adaptation has been studied extensively for mitigating data annotation constraints. Most existing works can be broadly classified into three categories. The first category is \textit{adversarial training} based which employs a discriminator to align source and target domains in the feature, output or latent space \cite{long2016unsupervised,tzeng2017adversarial,luo2019taking,tsai2018learning,saito2018maximum,huang2021mlan,vu2019advent,tsai2019domain,huang2021category,yang2020part,zhang2021detr,guan2021uncertainty,huang2021semi}. The second category is \textit{image translation} based which adapt image appearance to mitigate domain gaps \cite{li2019bidirectional,zhan2019ga,yang2020fda,zhang2021spectral}. The third category is \textit{self-training} based which predict pseudo labels or minimize entropy to guide iterative learning from target samples \cite{zou2018unsupervised, zou2019confidence,luo2021unsupervised,guan2021scale,huang2021cross,guan2021domain}.

Domain adaptation involves two typical training losses, namely, supervised loss over labeled source data and unsupervised loss over unlabeled target data. State-of-the-art methods tend to over-minimize the two types of losses which directly leads to deviated models with suboptimal adaptation as illustrated in Fig.~\ref{fig:intro}. We design a robust domain adaptation technique that addresses this issue by preventing loss over-minimization.

\textbf{Overfitting in Network Training:} Overfitting is a common phenomenon in deep network training which has been widely studied in the deep learning and computer vision community~\cite{belkin2018overfitting,caruana2001overfitting,ng1997preventing,roelofs2019meta,werpachowski2019detecting,wang2021embracing}. Most existing works address overfitting by certain regularization strategies such as weight decay~\cite{hanson1988comparing}, dropout~\cite{srivastava2014dropout}, $l_{1}$ regularization~\cite{tibshirani1996regression}, mix-up~\cite{zhang2017mixup}, label smoothing~\cite{szegedy2016rethinking}, batch normalization~\cite{ioffe2015batch}, virtual adversarial training~\cite{miyato2018virtual}, flooding~\cite{ishida2020we}, etc. However, these strategies were mostly designed for supervised or semi-supervised learning which do not fit in well for domain adaptive learning that usually involves domain gaps and unsupervised losses. We design a adversarial attacking technique that mitigates the overfitting in domain adaptive learning effectively.

\textbf{Adversarial Attacking:} Adversarial attacking has been studied in various security problems. For example, \cite{szegedy2013intriguing} shows that adversarial samples
can easily confuse CNN models. The following works improved adversarial attacking from different aspects via fast gradient signs \cite{goodfellow2014explaining}, minimal adversarial perturbation \cite{moosavi2016deepfool}, universal adversarial perturbations \cite{moosavi2017universal}, gradient-free attacking \cite{baluja2017adversarial}, transferable adversarial sample generation\cite{liu2016delving}, etc. Adversarial attacking has also been applied to other tasks, e.g., \cite{zheng2016improving,miyato2018virtual} employed adversarial samples to mitigate over-fitting in supervised and semi-supervised learning, \cite{volpi2018generalizing} generated adversarial samples for data augmentation, and \cite{liu2019transferable, yang2020adversarial} augments transferable features for domain divergence minimization. 

Most existing adversarial attacking methods commonly constrain the magnitude of perturbation noises for minimal modification of image semantics. However, such generated adversarial samples cannot tackle overfitting in domain adaptive learning well which usually involves a domain gap of fair magnitude. We design an innovative Fourier adversarial attacking technique that allows to generate adversarial samples without magnitude constraint yet with minimal modification of image semantics, more details to be described in the ensuing subsections.


\section{Method}
\begin{figure*}[ht]
\centering
\subfigure{\includegraphics[width=0.98\linewidth]{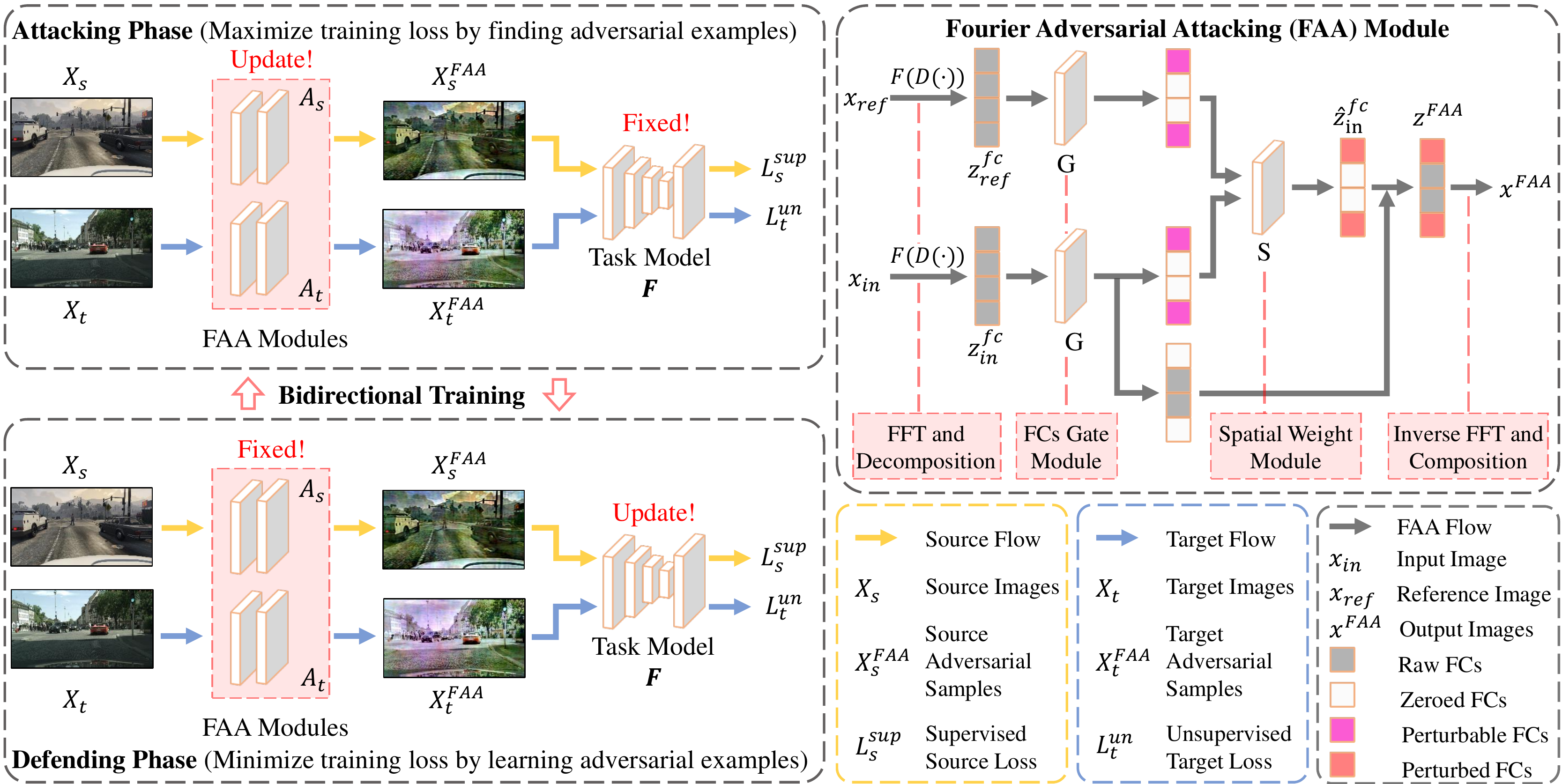}}
\caption{
Overview of our proposed robust domain adaptaion (RDA) via Fourier Adversarial Attacking (FAA). RDA has a bidirectional training process which consists of an \textit{Attacking Phase} and a \textit{Defending Phase}. Given a training image, the attacking phase learns to generate adversarial samples with minimal modification of image semantics. During the defending phase, the generated adversarial samples are applied to mitigate overfitting by preventing over-minimization of training losses. Given an input image, FAA decomposes it into multiple FCs representation, identifies perturbable FCs, and then perturbs them through weighted sum with the corresponding FCs from a randomly picked reference image. The objective of FFA is to maximize task losses while ensuring minimal modification of image semantics.
}
\label{fig:stru}
\end{figure*}

We achieve robust domain adaptation via Fourier adversarial attacking as illustrated in Fig.~\ref{fig:stru}. The training consists of two phases, namely, an \textit{Attacking Phase} and a \textit{Defending Phase}. Given a training image, the attacking phase learns to identify the right FCs with limited semantic information that allow perturbation noises of large magnitude. It also learns to generate adversarial samples (with perturbable FCs) with minimal modification of image semantics. During the defending phase, the generated adversarial sample is applied to mitigate overfitting by preventing over-minimization of training losses, more details to be described in the ensuing subsections.

\subsection{Task Definition}
We focus on the problem of unsupervised domain adaptation (UDA). Given labeled source data \{$X_{s}$, $Y_{s}$\} and unlabeled target data $X_{t}$, our goal is to learn a task model $F$ that performs well on $X_{t}$. The \textit{baseline} model is trained with the labeled source data only:
\begin{equation}
\begin{split}
\mathcal{L}_{sup}(X_{s}, Y_{s}; F) = l(F(X_{s}), Y_{s}),    
\end{split}
\label{eq_baseline}
\end{equation}
where $l(\cdot)$ denotes an accuracy-related loss, \eg, the standard cross-entropy loss.

\subsection{Fourier Adversarial Attacking}
Our proposed Fourier adversarial attacking (FAA) generates adversarial samples to attack the training loss to mitigate overfitting in domain adaptation, as shown in Fig.~\ref{fig:stru}. In adversarial sample generation, it first employs Fourier transformation to decompose input images into multiple frequency components (FCs) and then inject perturbation to non-semantic FCs which allows perturbation of large magnitude but with minimal modification of image semantics.

\textbf{Fourier Decomposition:} Inspired by JPEG \cite{wallace1992jpeg,pennebaker1992jpeg} and frequency-domain learning \cite{huang2021fsdr,jiang2020focal}, we transform an image $x$ into frequency space and decompose it into multiple FCs which allows explicit manipulation of each FC and more controllable perturbations. We employ Fourier transformation to convert $x$ into frequency space and further decompose it into multiple FCs of equivalent bandwidth:
\begin{equation}
\begin{split}
z = \mathcal{F}(x),\nonumber\\
z^{fc} = \mathcal{D}(z;N),\nonumber
\end{split}
\end{equation}
where $\mathcal{F}(\cdot)$ stands for Fourier transformation \cite{bracewell1986fourier}; $z$ denotes the frequency-space representation of $x$; $\mathcal{D}(z;N)$ denotes a function that decomposes $z$ into $N$ FCs $z^{fc} = \{z^{1},  z^{2}, ... , z^{N-1}, z^{N}\}$ of equivalent bandwidth.

We consider a gray-scale image $x \in \mathbb{R}^{H \times H}$ for defining $\mathcal{D}(z;N)$ formally. We thus have $z \in \mathbb{C}^{H \times H}$, where $\mathbb{C}$ denotes complex numbers. We use $z(i,j)$ to index $z$ at $(i,j)$, and $c_{i}$ and $c_{j}$ to denote the image centroid (\ie, image center at $(H/2, H/2)$). The equation $\{z^{1},  z^{2}, ... , z^{N-1}, z^{N}\}= \mathcal{D}(z;N)$ can be formally defined by:
\begin{equation}
z^{n}(i, j) =
\left\{
\begin{array}{ll}
z(i, j), &\text{if} \ \frac{n-1}{N} < d((i,j),(c_{i},c_{j})) < \frac{n}{N},\\
0, &\text{otherwise},\\
\end{array}
\right.\nonumber
\end{equation}
where $d(\cdot,\cdot)$ denotes Euclidean distance, $N$ denotes how many FCs the input is supposed to be decomposed into and $n$ denotes the FC index. 

To get $z^{fc} = \{z^{1},  z^{2}, ... , z^{N-1}, z^{N}\}= \mathcal{D}(z;N)$, we perform the equation $N$ times by changing $n$ from $1$ to $N$. If $x$ does not have a square size, we first up-sample the short side to be same as the long side before this process and down-sample it back to the original size after processing. If $x$ has more than one channel ($e.g.$, RGB image $x \subset \mathbb{R}^{n \times n \times 3}$), we process each channel independently.

\textbf{Adversarial Attacking:} With the decomposed FCs $z^{fc} = \{z^{1},  z^{2}, ... , z^{N-1}, z^{N}\}$, we attack domain adaption losses by perturbing partial FCs without magnitude constraint. Specifically, we employ a learnable gate module \cite{jang2016categorical,tucker2017rebar, xu2020learning,maddison2016concrete} to select certain FCs for perturbation. This gate $G$ assigns a binary score to each FC, where `1' indicates this FC is selected for perturbation (multiplied by `1') while `0' shows this FC is discarded (reset to all zero values). $G$ works with Gumbel-Softmax, a differentiable sampling mechanism for categorical variables that can be trained via standard back-propagation. Please refer to \cite{jang2016categorical, tucker2017rebar, xu2020learning, maddison2016concrete} for details.

Given a reference image (an randomly selected target-domain image) in FC representation $z^{fc}_{ref} = \{z^{1}_{ref},  z^{2}_{ref}, ... , z^{N-1}_{ref}, z^{N}_{ref}\}= \mathcal{D}(\mathcal{F}(x_{ref});N)$, we employ this gate module to block some FCs (all reset to zero), and apply the selected FCs to perturb the corresponding FCs of the input image $x$:
\begin{equation}
\hat{z}^{fc} = (1 - G(z^{fc}))z^{fc} + G(z^{fc})z^{fc}_{ref},
\label{eq_gating}
\end{equation}
where the learnable gate module $G$ enables binary channel attention that selects which FCs to perturb.

Specifically, for the identified FCs of the input image $x$, we extract the corresponding FCs of a reference image $X_{ref}$ and use them as perturbation noises. We generate adversarial samples in this manner because the perturbation noises (\ie, non-semantic FCs of $z^{fc}_{ref}$) from target natural images are more reasonable and meaningful as compared to a random noises/signals as in many existing adversarial sample generation methods. In addition, the use of non-semantic FCs of target samples mitigates inter-domain gaps which helps to improve target-domain performance in domain adaptation.

With the perturbed FCs $\hat{z}^{fc}$, we convert them all back (via inverse Fourier transformation) to get the adversarial sample $x^{FAA}$:
\begin{equation}
x^{FAA} = \mathcal{F}^{-1}( \mathcal{C}(\hat{z}^{fc})),
\end{equation}
where $\mathcal{C}(\cdot)$ denotes an inverse process of $\mathcal{D}(\cdot;N)$ that `compose’ the decomposed $z^{fc}$ back to the full representation by summing all the elements over frequency channels.

The \textit{Fourier Decomposition} and \textit{Adversarial Attacking} can be combined to form the FAA as follows:
\begin{equation}
x^{FAA} = \mathcal{A}(x),
\label{eq_attacker}
\end{equation}
where $\mathcal{A}$ takes an image $x$ and outputs an adversarial sample $x^{FAA}$ via the FAA as described. $\mathcal{A}$ has two sub-modules with learnable parameters, $i.e.$, the gate module $G$ and the single-layer neural network for spatial weighting map $\mathcal{S}$ generation. The rest operations in $\mathcal{A}$ are deterministic such as Fourier transformation and its inverse ($\mathcal{F}(\cdot)$ and $\mathcal{F}^{-1}(\cdot)$), decomposition and re-composition ($\mathcal{D}(\cdot;N)$ and $ \mathcal{C}(\cdot)$)

\subsection{FAA Training}
The proposed FAA involves three types of losses including task loss (\ie, UDA training loss in this work) that is to be attacked, the gate related loss that constrains the proportion of perturbable FCs, and the reconstruction loss that aims to minimize the attacking effects over image semantics. 

Given input images $X \subset \mathbb{R}^{H \times W \times 3}$, a task model $F$ and the Fourier attacker $\mathcal{A}$, we denote the task loss by $\mathcal{L}(X;\mathcal{A},F)$. Note the task loss refers to UDA training loss in this work but it can be other types of losses such as supervised or unsupervised training losses. The gate related loss and the reconstruction loss losses are defined by:
\begin{equation}
\begin{split}
\mathcal{L}_{gat}(X;\mathcal{A}) =
\left\{
\begin{array}{ll}
\sum(G(z^{fc})), &\text{if} \ \sum(G(z^{fc}))>N*p,\\
0, &\text{otherwise},\\
\end{array}
\right.\\\nonumber
\mathcal{L}_{rec}(X;\mathcal{A}) = |R(X)- R(\mathcal{A}(X))|,\nonumber
\end{split}
\label{eq_gate_recon}
\end{equation}
where $G(z^{fc})$ is a gating process in $\mathcal{A}$ as described in Eq.~\ref{eq_gating}. $R$ is a band-pass filter to get the mid-frequency content \cite{huang2021fsdr} that captures semantic information ($e.g.$, structures and shapes) and thus the consistency loss can ensure that the selected FCs contains minimal semantic information. $p$ is a hyper-parameter that constrains the maximum number of perturbable FCs by $N*p$.

The overall training objective of FAA is formulated by:
\begin{equation}
\max_{\mathcal{A}}  \mathcal{L}(X;\mathcal{A},F) - \mathcal{L}_{gat}(X;\mathcal{A}) - \mathcal{L}_{rec}(X;\mathcal{A})
\label{eq_attacking}
\end{equation}

\subsection{Robust Domain Adaptation}
Sections 3.2 and 3.3 describe the ``Attacking Phase" that generates adversarial samples via FAA.
This section presents the proposed robust domain adaptation technique where FAA is used to mitigate the overfitting in UDA in a ``Defending Phase”. Specifically, we apply the FAA-generated adversarial samples to prevent the over-minimization of UDA training losses in domain adaptation.

Given the task model $F$ and the Fourier attacker $\mathcal{A}$, the training objective of the task model $F$ is formulated by:
\begin{equation}
\min_{F}  \mathcal{L}(X;\mathcal{A},F)
\label{eq_defending}
\end{equation}

The optimization functions of both attacking (\ie, Eq.~\ref{eq_attacking}) and defending (\ie, Eq.~\ref{eq_defending}) are generic and applicable to various \textit{tasks} and \textit{data}. Specifically, the model $F$ could be for semantic segmentation, object detection or image classification task. The data could be labeled source data with supervised source losses or unlabeled target data with unsupervised target losses such as adversarial loss~\cite{tsai2018learning, luo2019taking, vu2019advent, yang2020adversarial, Pan_2020_CVPR, liu2019transferable, huang2020contextual, saito2019strong}, self-training loss~\cite{zou2018unsupervised, zou2019confidence, li2019bidirectional, wang2020differential, kim2020learning,yang2020fda} or entropy loss~\cite{vu2019advent, grandvalet2005semi}, etc. Below are a supervised loss and a self-training-based unsupervised loss here for reference:
\begin{equation}
\begin{split}
\mathcal{L}_{s}^{sup}(X_{s};\mathcal{A}, F) = l(F(\mathcal{A}(X_{s})), Y_{s}), \\\nonumber
\mathcal{L}_{t}^{un}(X_{t};\mathcal{A}, F) = l(F(\mathcal{A}(X_{t})), \hat{Y}_{t}), \nonumber
\end{split}
\label{eq_gate_recon}
\end{equation}
where $l(\cdot)$ denotes an accuracy-related loss, \eg, cross entropy loss; $\mathcal{A}(X_{s})$ and $\mathcal{A}(X_{t})$ are inputs perturbed by FAA, and $\hat{Y}_{t}$ is the pseudo label of unlabeled target data.

\begin{algorithm}[t]
    \caption{The proposed Robust Domain Adaptation via Fourier adversarial attacking (FAA).
    }\label{algorithm_FAA_RDA}
    \begin{algorithmic}[1]
        \REQUIRE Input data $X$; task model $F$;
        Fourier attacker $\mathcal{A}$
        \ENSURE Learnt parameters $\theta$ of task model $F$
        \FOR{$iter = 1$ \textbf{to} $Max\_Iter$}
            \STATE Sample a batch of $X$ uniformly from the dataset
                \STATE \textbf{\emph{Defending phase (minimize loss):}}
                \STATE Update $F$ via Eq.~\ref{eq_defending}
                \STATE \textbf{\emph{Attacking phase (maximize loss):}}
                \STATE Update $\mathcal{A}$ via Eq.~\ref{eq_attacking}
        \ENDFOR
  \RETURN $F$
  \end{algorithmic}
\end{algorithm}

In summary, our robust domain adaptation has a bidirectional training framework including an \textit{Attacking Phase} and a \textit{Defending Phase} as shown in Fig.~\ref{fig:stru}. In the \textit{Attacking Phase}, the task model $F$ is fixed and the attacker $\mathcal{A}$ generates adversarial samples to increase the training loss. In the \textit{Defending Phase}, $\mathcal{A}$ is fixed and $F$ is updated to reduce the training loss. These two phases are conducted in an alternative way way and form a bidirectional training framework (adversarial training). Please refer to Algorithm~\ref{algorithm_FAA_RDA} for details.

\begin{table*}[!t]
	\centering
	\resizebox{\linewidth}{!}{
	\begin{tabular}{c|ccccccccccccccccccc|c}
		\hline
		Method  & Road & SW & Build & Wall & Fence & Pole & TL & TS & Veg. & Terrain & Sky & PR & Rider & Car & Truck & Bus & Train & Motor & Bike & mIoU\\
		\hline
		Baseline~\cite{he2016deep} &75.8	&16.8	&77.2	&12.5	&21.0	&25.5	&30.1	&20.1	&81.3	&24.6	&70.3	&53.8	&26.4	&49.9	&17.2	&25.9	&6.5	&25.3	&36.0	&36.6\\
		\textbf{+FAA-S}  &89.8	&39.1	&81.7	&27.6	&19.9	&34.2	&35.9	&23.3	&82.1	&29.5	&76.6	&58.3	&26.0	&82.1	&32.5	&45.2	&15.3	&26.9	&33.5	&\textbf{45.2}\\\hline
		AdaptSeg~\cite{tsai2018learning}  &86.5	&36.0	&79.9	&23.4	&23.3	&23.9	&35.2	&14.8	&83.4	&33.3	&75.6	&58.5	&27.6	&73.7	&32.5	&35.4	&3.9	&30.1	&28.1	&42.4\\
		\textbf{+FAA-S}  &89.1	&38.7	&82.0	&28.5	&23.6	&33.2	&34.9	&22.2	&83.7	&34.0	&77.5	&59.0	&26.6	&83.5	&33.8	&45.8	&15.8	&28.3	&33.9	&46.0\\
		\textbf{+FAA-T}  &91.0	&44.3	&82.3	&28.1	&21.7	&35.6	&36.8	&18.7	&84.2	&38.3	&73.7	&61.4	&28.5	&85.9	&38.6	&45.7	&14.9	&32.4	&33.3	&47.1\\
		\textbf{+FAA}  &91.2	&44.7	&82.5	&27.7	&24.7	&36.7	&36.5	&24.7	&84.9	&38.1	&78.4	&62.2	&27.4	&84.9	&39.3	&46.7	&12.9	&31.9	&36.7	&\textbf{48.0}\\\hline
		ST~\cite{zou2018unsupervised} &88.5	&26.3	&82.0	&24.5	&19.9	&33.6	&37.6	&19.8	&82.7	&26.2	&76.8	&60.0	&26.6	&82.9	&28.9	&31.0	&6.3	&24.5	&31.2	&42.6\\
		\textbf{+FAA-S}  &91.4	&52.1	&81.6	&32.6	&26.4	&35.3	&36.2	&22.5	&84.6	&35.8	&77.3	&60.5	&27.9	&84.1	&37.0	&43.8	&17.2	&25.6	&35.9	&47.8\\
		\textbf{+FAA-T}  &91.6	&49.0	&83.2	&32.7	&26.5	&38.9	&46.5	&23.6	&83.2	&35.3	&77.9	&60.1	&28.5	&85.5	&38.1	&41.4	&16.9	&25.6	&36.4	&48.5\\
		\textbf{+FAA}  &92.0	&52.6	&82.5	&33.9	&28.5	&39.4	&45.7	&33.6	&84.3	&38.4	&80.5	&60.4	&29.1	&85.4	&39.9	&44.4	&16.2	&29.6	&35.1	&\textbf{50.1}\\
		\hline
        CLAN~\cite{luo2019taking}  &87.0 &27.1 &79.6 &27.3 &23.3 &28.3 &35.5 &24.2 &83.6 &27.4 &74.2 &58.6 &28.0 &76.2 &33.1 &36.7 &{6.7} &{31.9} &31.4 &43.2\\
        \textbf{+FAA}  &92.0	&53.4	&82.8	&32.5	&29.8	&36.8	&36.6	&28.5	&83.3	&35.8	&77.1	&61.2	&30.1	&83.8	&36.8	&46.2	&11.7	&28.6	&35.6	&\textbf{48.6}\\\hline
        AdvEnt~\cite{vu2019advent}  &89.4 &33.1 &81.0 &26.6 &26.8 &27.2 &33.5 &24.7 &{83.9} &{36.7} &78.8 &58.7 &30.5 &{84.8} &38.5 &44.5 &1.7 &31.6 &32.4 &45.5\\
        \textbf{+FAA}  &92.1	&49.0	&82.5	&31.2	&28.6	&38.3	&37.1	&27.1	&84.8	&39.4	&79.1	&61.3	&29.1	&85.0	&38.2	&46.6	&14.3	&32.4	&36.9	&\textbf{49.1}\\\hline
        IDA~\cite{pan2020unsupervised}  &90.6 &37.1 &82.6 &30.1 &19.1 &29.5 &32.4 &20.6 &85.7 &40.5 &79.7 &58.7 &31.1 &86.3 &31.5 &48.3 &0.0 &30.2 &35.8 &46.3\\
        \textbf{+FAA}  &91.4	&50.3	&83.4	&33.0	&27.2	&37.7	&38.4	&27.0	&84.4	&41.8	&79.9	&59.7	&30.6	&85.1	&37.7	&47.5	&14.2	&31.2	&35.3	&\textbf{49.3}\\\hline
        CRST~\cite{zou2019confidence}  &91.0 &55.4 &80.0 &33.7 &21.4 &37.3 &32.9 &24.5 &85.0 &34.1 &80.8 &57.7 &24.6 &84.1 &27.8 &30.1 &26.9 &26.0 &42.3 &47.1\\
        \textbf{+FAA}  &92.7	&56.4	&84.4	&34.3	&29.2	&38.2	&48.8	&47.1	&85.5	&42.4	&86.0	&62.3	&32.9	&85.3	&40.8	&49.8	&22.7	&23.0	&36.8	&\textbf{52.6}\\\hline
        BDL~\cite{li2019bidirectional}  &91.0  &44.7  &84.2  &34.6  &27.6 &30.2  &36.0  &36.0 &85.0  &43.6  &83.0 &58.6 &31.6  &83.3  &35.3  &49.7 &3.3  &28.8  &35.6 &48.5\\
        \textbf{+FAA}  &92.8	&54.1	&85.3	&33.8	&28.1	&41.0	&46.1	&47.6	&84.8	&42.7	&82.8	&63.5	&32.5	&84.1	&37.0	&50.3	&15.5	&23.2	&36.5	&\textbf{51.7}\\\hline
        CrCDA~\cite{huang2020contextual}  &92.4	&55.3	&82.3	&31.2	&29.1	&32.5	&33.2	&35.6	&83.5	&34.8	&84.2	&58.9	&32.2	&84.7	&40.6	&46.1	&2.1	&31.1	&32.7	&48.6 \\
        \textbf{+FAA}  &92.9	&55.2	&83.4	&31.8	&31.1	&39.4	&47.8	&46.4	&83.4	&38.4	&85.7	&61.2	&31.0	&85.7	&39.1	&46.3	&12.0	&33.4	&38.9	&\textbf{51.7}\\\hline
        SIM~\cite{wang2020differential}  &90.6 &44.7 &84.8 &34.3 &28.7 &31.6 &35.0 &37.6 &84.7 &43.3 &85.3 &57.0 &31.5 &83.8 &42.6 &48.5 &1.9 &30.4 &39.0 &49.2\\
        \textbf{+FAA}  &92.2	&53.7	&84.6	&34.2	&29.1	&38.0	&47.0	&45.3	&85.0	&43.8	&84.9	&60.3	&29.2	&85.2	&39.5	&47.3	&12.4	&32.3	&44.3	&\textbf{52.0}\\\hline
        TIR~\cite{kim2020learning}  &92.9 &55.0 &85.3 &34.2 &31.1 &34.9 &40.7 &34.0 &85.2 &40.1 &87.1 &61.0 &31.1 &82.5 &32.3 &42.9 &0.3 &36.4 &46.1 &50.2\\
        \textbf{+FAA}  &93.0	&55.5	&84.0	&33.0	&28.2	&38.1	&46.6	&45.8	&84.9	&41.7	&86.1	&61.2	&33.7	&84.2	&38.3	&46.5	&15.3	&34.9	&44.6	&\textbf{52.4}\\\hline
        FDA~\cite{yang2020fda}  &92.5 &53.3 &82.4 &26.5 &27.6 &36.4 &40.6 &38.9 &82.3 &39.8 &78.0 &62.6 &34.4 &84.9 &34.1 &53.1 &16.9 &27.7 &46.4 &50.5\\
        \textbf{+FAA}  &93.1	&55.0	&84.7	&33.1	&29.5	&38.7	&49.3	&44.9	&84.8	&41.6	&80.2	&62.3	&33.2	&85.6	&37.3	&51.3	&18.5	&34.6	&45.3	&\textbf{52.8}\\\hline
	\end{tabular}
	}
	\caption{
	Domain adaptive semantic segmentation experiments over task GTA5 $\rightarrow$ Cityscapes (\textit{+FAA-S}, \textit{+FAA-T}, and \textit{+FAA} mean to apply our FAA to attack supervised source losses, unsupervised target losses, and both types of losses, respectively).}
	\label{table:gta2city}
\end{table*}

\begin{table*}[h]
	\centering
	\resizebox{\linewidth}{!}{
	\begin{tabular}{c|cccccccccccccccc|c|c}
		\hline
		Method  & Road & SW & Build & Wall\textsuperscript{*} & Fence\textsuperscript{*} & Pole\textsuperscript{*} & TL & TS & Veg. & Sky & PR & Rider & Car & Bus & Motor & Bike & mIoU & mIoU\textsuperscript{*}\\
		\hline
		Baseline~\cite{he2016deep} &55.6	&23.8	&74.6	&9.2	&0.2	&24.4	&6.1	&12.1	&74.8	&79.0	&55.3	&19.1	&39.6	&23.3	&13.7	&25.0	&33.5	&38.6\\
		PatAlign~\cite{tsai2019domain}  &82.4 &38.0 &78.6 &8.7 &0.6 &26.0 &3.9 &11.1 &75.5 &84.6 &53.5 &21.6 &71.4 &32.6 &19.3 &31.7 &40.0 &46.5\\
        AdaptSeg~\cite{tsai2018learning}  &84.3 &42.7 &77.5 &- &- &- &4.7 &7.0 &77.9 &82.5 &54.3 &21.0 &72.3 &32.2 &18.9 &32.3 &- &46.7\\
        CLAN~\cite{luo2019taking}  &81.3 &37.0 &{80.1} &- &- &- &{16.1} &{13.7} &78.2 &81.5 &53.4 &21.2 &73.0 &32.9 &{22.6} &30.7 &- &47.8\\
        AdvEnt~\cite{vu2019advent}  &85.6 &42.2 &79.7 &{8.7} &0.4 &25.9 &5.4 &8.1 &{80.4} &84.1 &{57.9} &23.8 &73.3 &36.4 &14.2 &{33.0} &41.2 &48.0\\
        IDA~\cite{pan2020unsupervised}  &84.3 &37.7 &79.5 &5.3 &0.4 &24.9 &9.2 &8.4 &80.0 &84.1 &57.2 &23.0 &78.0 &38.1 &20.3 &36.5 &41.7 &48.9\\
        CrCDA~\cite{huang2020contextual}   &86.2	&44.9	&79.5	&8.3	&0.7	&27.8	&9.4	&11.8	&78.6	&86.5	&57.2	&26.1	&76.8	&39.9	&21.5	&32.1	&42.9	&50.0\\\hline
        TIR~\cite{kim2020learning}  &92.6 &53.2 &79.2 &- &- &- &1.6 &7.5 &78.6 &84.4 &52.6 &20.0 &82.1 &34.8 &14.6 &39.4 &- &49.3\\
        \textbf{+FAA} &89.1	&51.3	&78.7	&8.9	&0.1	&27.1	&19.1	&17.0	&82.4	&83.9	&55.8	&27.7	&84.5	&39.6	&26.9	&38.3	&45.7	&\textbf{53.4}\\\hline
        CRST~\cite{zou2019confidence}  &67.7 &32.2 &73.9 &10.7 &1.6 &37.4 &22.2 &31.2 &80.8 &80.5 &60.8 &29.1 &82.8 &25.0 &19.4 &45.3 &43.8 &50.1\\
        \textbf{+FAA} &89.3	&49.6	&79.3	&12.9	&0.1	&37.2	&20.6	&29.8	&83.7	&84.0	&62.6	&28.0	&84.9	&37.1	&30.1	&48.2	&48.6	&\textbf{55.9}\\\hline
        BDL~\cite{li2019bidirectional}  &86.0   &46.7   &80.3&-&-&-  &14.1   &11.6 &79.2 &81.3 &54.1   &27.9   &73.7   &42.2   &25.7   &45.3  &- &51.4\\
        \textbf{+FAA} &89.7	&47.5	&82.7	&6.2	&0.1	&31.8	&25.0	&22.7	&82.3	&84.8	&59.0	&26.2	&83.4	&39.2	&31.5	&37.9	&46.9	&\textbf{54.8}\\\hline
        SIM~\cite{wang2020differential}  &83.0 &44.0 &80.3 &- &- &- &17.1 &15.8 &80.5 &81.8 &59.9 &33.1 &70.2 &37.3 &28.5 &45.8 &- &52.1\\
        \textbf{+FAA} &88.3	&47.8	&82.8	&9.3	&1.1	&36.7	&24.5	&26.8	&79.4	&83.7	&62.5	&32.1	&81.7	&36.3	&29.2	&46.4	&48.0	&\textbf{55.5}\\\hline
        FDA~\cite{yang2020fda}  &79.3 &35.0 &73.2 &- &- &- &19.9 &24.0 &61.7 &82.6 &61.4 &31.1 &83.9 &40.8 &38.4 &51.1 &- &52.5\\
        \textbf{+FAA} &89.4	&39.0	&79.8	&8.1	&1.2	&33.0	&22.6	&28.1	&81.8	&82.0	&60.9	&30.1	&82.7	&41.4	&37.5	&48.9	&47.9	&\textbf{55.7}\\\hline
	\end{tabular}
	}
	\caption{
	Domain adaptive semantic segmentation experiments over task SYNTHIA $\rightarrow$ Cityscapes (\textit{+FAA} means to include FAA to attack supervised source loss and unsupervised target loss in domain adaptation).}
	\label{table:synthia2city}
\end{table*}

This bidirectional training framework uses FAA to prevent over-minimization of UDA losses by forcing them oscillating around a small value. In another word, it ensures that the task model $F$ can continue the “random walk” and drift into an area with a flat loss landscape~\cite{chaudhari2019entropy,keskar2016large,li2017visualizing}, leading to robust domain adaptation.

\begin{table*}[t]
\centering
\footnotesize
\begin{tabular}{l|p{1cm}<{\centering}p{1cm}<{\centering}p{1cm}<{\centering}p{1cm}<{\centering}p{1cm}<{\centering}
p{1cm}<{\centering}p{1cm}<{\centering}p{1cm}<{\centering}|c}
\hline
Method & person & rider & car & truck & bus & train & mcycle & bicycle & mAP \\
\hline
Baseline~\cite{ren2015faster} &  24.4 & 30.5  & 32.6 & 10.8  & 25.4 & 9.1 & 15.2 & 28.3 & 22.0 \\
MAF~\cite{he2019multi} &  28.4 & 39.5  & 43.9 & 23.8  & 39.9 & 33.3 & 29.2 & 33.9 & 34.0 \\
SCDA~\cite{zhu2019adapting} &  33.5 & 38.0  & 48.5 & 26.5  & 39.0 & 23.3 & 28.0 & 33.6 & 33.8 \\
DA~\cite{chen2018domain} &25.0 &31.0 &40.5 &22.1 &35.3 &20.2 &20.0 &27.1 &27.6\\
MLDA~\cite{Xie_2019_ICCV} &33.2 &44.2 &44.8 &28.2  &41.8  &28.7 &30.5 &36.5 &36.0\\
DMA~\cite{kim2019diversify} &30.8 &40.5 &44.3 &27.2 &38.4 &34.5 &28.4 &32.2 &34.6\\
CAFA~\cite{hsu2020every} &41.9 &38.7 &56.7 &22.6 &41.5 &26.8 &24.6 &35.5 &36.0 \\
\hline
SWDA~\cite{saito2019strong} &36.2 &35.3   &43.5 &30.0 &29.9 &42.3  &32.6 &24.5 &34.3\\
\textbf{+FAA} &39.5	&41.3	&47.0	&34.5	&39.3	&44.0	&31.9	&28.4	&\textbf{38.3}\\\hline
CRDA~\cite{xu2020exploring} & 32.9 & 43.8  & 49.2 & 27.2  & 45.1 & 36.4 & 30.3 & 34.6 & 37.4 \\
\textbf{+FAA} &37.4	&46.4	&48.3	&32.6	&46.5	&39.3	&32.3	&35.3	&\textbf{39.8}\\
\hline
\end{tabular}
\caption{Experimental results of domain adaptive object detection over the adaptation task Cityscapes $\rightarrow$ Foggy Cityscapes (\textit{+FAA} means to include FAA to attack supervised source loss and unsupervised target loss in domain adaptation).}
\label{table:det_city2fog}
\end{table*}

\section{Experiments}
This section presents experiments including datasets and implementation details, domain adaptation for semantic segmentation (with ablation studies), object detection, and image classification tasks, and discussion, more details to be described in the ensuing subsections.

\subsection{Datasets}
\textbf{Adaptation for semantic segmentation:} We consider two synthesized-to-real segmentation tasks: 1) GTA5 \cite{richter2016playing} $\rightarrow$ Cityscapes \cite{cordts2016cityscapes} and 2) SYNTHIA \cite{ros2016synthia} $\rightarrow$ Cityscapes. GTA5 contains $24,966$ synthetic images and shares $19$ categories with Cityscapes. For SYNTHIA, we use `SYNTHIA-RAND-CITYSCAPES' which contains $9,400$ synthetic images and shares 16 categories with Cityscapes. Cityscapes contains $2975$/$500$ training/validation images. Following \cite{tsai2018learning, zou2018unsupervised}, we adapt towards the Cityscapes training set and evaluate on the Cityscapes validation set.

\textbf{Adaptation for object detection:} We consider two adaptation tasks: 1) Cityscapes $\rightarrow$ Foggy Cityscapes~\cite{sakaridis2018semantic} and 2) Cityscapes $\rightarrow$ BDD100k~\cite{yu2018bdd100k}. For Cityscapes, we convert instance segmentation annotations to bounding boxes in experiments. Foggy Cityscapes was created by applying synthetic fog on Cityscapes images. BDD100k contains $100k$ images including $70k$ for training and $10k$ for validation. Following \cite{xu2020exploring,saito2019strong,chen2018domain}, we use a BDD100k subset \textit{daytime} in experiments, which includes $36,728$ training images and $5,258$ validation images.

\begin{table*}[h]
\centering
\footnotesize
\begin{tabular}{l|p{1cm}<{\centering}p{1cm}<{\centering}p{1cm}<{\centering}p{1cm}<{\centering}p{1cm}<{\centering}p{1cm}<{\centering}p{1cm}<{\centering}|c} 
\hline
Method & person & rider & car & truck & bus & mcycle & bicycle & mAP \\
\hline
Baseline~\cite{ren2015faster} &  26.9 & 22.1 & 44.7 & 17.4  & 16.7 & 17.1 & 18.8 & 23.4 \\
DA~\cite{chen2018domain} & 29.4 & 26.5  & 44.6 & 14.3  & 16.8 & 15.8 & 20.6 & 24.0 \\
\hline
SWDA~\cite{saito2019strong} & 30.2 & 29.5  & 45.7 & 15.2  & 18.4 & 17.1 & 21.2 & 25.3 \\
\textbf{+FAA} &32.0	&32.2	&49.7	&19.8	&23.9	&18.5	&24.0	&\textbf{28.6}\\\hline
CRDA~\cite{xu2020exploring} & 31.4 & 31.3  & 46.3 & 19.5  & 18.9 & 17.3 & 23.8 & 26.9 \\
\textbf{+FAA} &32.9	&33.1	&51.4	&21.8	&24.5	&19.1	&26.2	&\textbf{29.9}\\
\hline
\end{tabular}
\caption{
Experimental results of domain adaptive object detection over the adaptation task Cityscapes $\rightarrow$ BDD100k (\textit{+FAA} means to include FAA to attack supervised source loss and unsupervised target loss in domain adaptation).}
\label{table:det_city2BDD}
\end{table*}

\textbf{Adaptation for image classification:} We adopt two adaptation benchmarks VisDA17~\cite{peng2018visda} and Office-31~\cite{saenko2010adapting}. VisDA17 includes $152,409$ synthetic images of $12$ categories as source and $55,400$ real images as target. Office-31 contains images of $31$ classes from Amazon (A), Webcam (W) and DSLR (D) that have $2817$, $795$ and $498$ images, respectively. We evaluate on six tasks A$\rightarrow$W, D$\rightarrow$W, W$\rightarrow$D, A$\rightarrow$D, D$\rightarrow$A, and W$\rightarrow$A as in~\cite{zou2019confidence, saenko2010adapting, sankaranarayanan2018generate}.

\subsection{Implementation Details}

\textbf{Semantic segmentation:} We use DeepLab-V2~\cite{chen2017deeplab} with ResNet101 \cite{he2016deep} as the segmentation network as in~\cite{tsai2018learning,zou2018unsupervised}. We use SGD optimizer \cite{bottou2010large} with a momentum $0.9$ and a weight decay $1e-4$. The initial learning rate is $2.5e-4$ and decayed by a polynomial policy of power $0.9$ \cite{chen2017deeplab}.

\textbf{Object detection:} As in~\cite{xu2020exploring, saito2019strong, chen2018domain}, we use Faster R-CNN~\cite{chen2017deeplab} with VGG-16~\cite{simonyan2014very} as the detection network. We use SGD optimizer \cite{bottou2010large} with a momentum $0.9$ and a weight decay $5e-4$. The initial learning rate is $1e-3$ for $50k$ iterations and reduced to $1e-4$ for another $20k$ iterations \cite{xu2020exploring,saito2019strong,chen2018domain}. In all experiments, we set image shorter side to 600 and employ RoIAlign~\cite{he2017mask} for feature extraction.

\textbf{Image classification:} For fair comparisons, we follow ~\cite{zou2019confidence, saenko2010adapting,sankaranarayanan2018generate} and use ResNet-101/ResNet-50~\cite{he2016deep} (pre-trained with ImageNet \cite{deng2009imagenet}) as backbones. We use SGD optimizer \cite{bottou2010large} with a momentum $0.9$ and a weight decay $5e-4$. The learning rate is $1e-3$ and the batch size is $32$ \cite{zou2019confidence}.

We set the parameter $p$ and the number of FCs $N$ at $0.1$ and $96$. 
The band-pass filter $R$ follows \cite{huang2021fsdr} with mid-pass and low-/high-rejected designs to get the mid-frequency content that captures semantic information (e.g., structures and shapes).

\subsection{Domain Adaptive Semantic Segmentation}

Table~\ref{table:gta2city} shows experimental results over semantic segmentation task GTA5 $\rightarrow$ Cityscapes. It can be seen that FAA is generic and can be applied to attack both \textit{Baseline} (for preventing overfitting in supervised source loss) and state-of-the-art UDA methods (for preventing overfitting in both supervised source loss and unsupervised target loss). In addition, incorporating FAA improves both \textit{Baseline} and UDA methods clearly and consistently. 

We perform ablation studies over two representative UDA methods using adversarial alignment~\cite{tsai2018learning} and self-training~\cite{zou2018unsupervised}, where \textit{+FAA-S}, \textit{+FAA-T} and \textit{+FAA} address the overfitting of supervised source losses, unsupervised target losses and both losses, respectively. It can be seen that \textit{+FAA-S} and \textit{+FAA-T} both improve domain adaptation by large margins. This shows that both supervised source objective and unsupervised target objective introduce clear overfitting and FAA mitigates the overfitting effectively. In addition, \textit{+FAA} performs clearly the best. This shows that preventing the two learning objectives from over-minimization is complementary as overftting in the two learning objectives affects generalization in different manners. Specifically, supervised source loss has domain gap and over-minimizing it guides the model to over-memorize source data whereas unsupervised target loss is noisy and over-minimizing it leads to deviated solutions with accumulated errors.

\begin{table*}[!t]
	\centering
	\centering
	\footnotesize
	\begin{tabular}{c|cccccccccccc|c}
		\hline
		Method & Aero & Bike & Bus & Car & Horse & Knife & Motor & Person & Plant & Skateboard & Train & Truck & Mean\\
		\hline
		Res-101 \cite{saito2018adversarial} & 55.1 & 53.3 & 61.9 & 59.1 & 80.6 & 17.9 & 79.7 & 31.2 & 81.0 & 26.5 & 73.5 & 8.5 & 52.4\\
		MMD \cite{long2015learning} & 87.1 & 63.0 & 76.5 & 42.0 & 90.3 & 42.9 & 85.9 & 53.1 & 49.7 & 36.3 & 85.8 & 20.7 & 61.1\\
		DANN \cite{ganin2016domain} & 81.9 & 77.7 & 82.8 & 44.3 & 81.2 & 29.5 & 65.1 & 28.6 & 51.9 & 54.6 & 82.8 & 7.8 & 57.4\\ 
		ENT \cite{grandvalet2005semi} & 80.3 & 75.5 & 75.8 & 48.3 & 77.9 & 27.3 & 69.7 & 40.2 & 46.5 & 46.6 & 79.3 & 16.0 & 57.0\\
		MCD \cite{saito2018maximum} & 87.0 & 60.9 & {83.7} & 64.0 & 88.9 & 79.6 & 84.7 & {76.9} & {88.6} & 40.3 & 83.0 & 25.8 & 71.9\\
		ADR \cite{saito2018adversarial} & 87.8 & 79.5 & {83.7} & 65.3 & {92.3} & 61.8 & {88.9} & 73.2 & 87.8 & 60.0 & {85.5} & {32.3} & 74.8\\  
		SimNet-Res152 \cite{pinheiro2018unsupervised} & {94.3} & 82.3 & 73.5 & 47.2 & 87.9 & 49.2 & 75.1 & 79.7 & 85.3 & 68.5 & 81.1 & 50.3 & 72.9\\
		GTA-Res152 \cite{sankaranarayanan2018generate} & - & - & - & - & - & - & - & - & - & - & - & - & 77.1\\
		\hline
		CBST~\cite{zou2018unsupervised} &87.2 & 78.8 & 56.5 & 55.4 & 85.1 & 79.2 & 83.8 &  77.7 & 82.8 & 88.8 & 69.0 & 72.0 & 76.4\\
        \textbf{CBST+FAA} &90.6	&80.3	&79.6	&67.1	&86.7	&80.2	&86.2	&77.1	&86.2	&87.1	&80.6	&71.8	&\textbf{81.1}\\\hline
		CRST~\cite{zou2019confidence} & 88.0 & 79.2 & 61.0 & 60.0 & 87.5 & 81.4 & 86.3 & 78.8 & 85.6 & 86.6 & 73.9 &   68.8 &78.1\\
        \textbf{CRST+FAA} &91.6	&80.5	&81.5	&70.7	&89.6	&81.0	&87.5	&79.9	&87.1	&86.4	&81.0	&75.1	&\textbf{82.7}\\
		\hline
	\end{tabular}
    \caption{
    Experimental results of domain adaptive image classification task on VisDA17 (\textit{+FAA} means to include FAA to attack supervised source loss and unsupervised target loss in domain adaptation).}
	\label{table:visda17}
\end{table*}

\begin{table}[t]
	\centering
	\resizebox{0.95\linewidth}{!}{
	\begin{tabular}{c|cccccc|c}
		\hline
		Method & A$\rightarrow$W & D$\rightarrow$W & W$\rightarrow$D & A$\rightarrow$D & D$\rightarrow$A & W$\rightarrow$A & Mean\\
		\hline
		ResNet-50 \cite{he2016deep} & 68.4 & 96.7 & 99.3 & 68.9 & 62.5 & 60.7 & 76.1\\
		DAN \cite{long2015learning} & 80.5 & 97.1 & 99.6 & 78.6 & 63.6 & 62.8 & 80.4\\
		RTN \cite{long2016unsupervised} & 84.5 & 96.8 & 99.4 & 77.5 & 66.2 & 64.8 & 81.6\\
		DANN \cite{ganin2016domain} & 82.0 & 96.9 & 99.1 & 79.7 & 68.2 & 67.4 & 82.2\\
		ADDA \cite{tzeng2017adversarial} & 86.2 & 96.2 & 98.4 & 77.8 & 69.5 & 68.9 & 82.9\\
		JAN \cite{long2017deep} & 85.4 & 97.4 & 99.8 & 84.7 & 68.6 & 70.0 & 84.3\\
		GTA \cite{sankaranarayanan2018generate} & {89.5} & 97.9 & 99.8 & 87.7 & 72.8 & 71.4 & 86.5\\
		\hline
		CBST~\cite{zou2018unsupervised} & 87.8 & 98.5 & {100} & 86.5 & 71.2 & 70.9 & 85.8\\
        \textbf{CBST+FAA}   &90.9   &98.9	&99.8	&92.5	&76.7	&76.2	&\textbf{89.1}\\\hline
		TAT~\cite{liu2019transferable} & 92.5 & 99.3 &100.0 & 93.2 & 73.1 & 72.1 & 88.4\\
		\textbf{TAT+FAA} &92.4	&99.8	&99.4	&94.7	&78.2	&77.1	&\textbf{90.3}\\\hline
		CRST~\cite{zou2019confidence} & 89.4 & {98.9} & {100} & 88.7 & 72.6 & 70.9 & 86.8\\
        \textbf{CRST+FAA} &92.3	&99.2	&99.7	&94.4	&80.5	&78.7	&\textbf{90.8}\\
		\hline
	\end{tabular}
	}
    \caption{
    Experimental results of domain adaptive image classification over Office-31 (\textit{+FAA} means to include FAA to attack supervised source loss and unsupervised target loss in domain adaptation).}
	\label{table:office}
\end{table}

Table~\ref{table:synthia2city} shows experimental results over semantic segmentation task SYNTHIA $\rightarrow$ Cityscapes. We can observe that FAA improves state-of-the-art UDA methods in the similar manner as in Table~\ref{table:gta2city}. Note we applied FAA to a few representative UDA methods only due to space limit.

\subsection{Domain Adaptive Object Detection}

Table~\ref{table:det_city2fog} shows domain adaptive object detection over the task Cityscapes $\rightarrow$ Foggy Cityscapes. It can be seen that FAA boosts mAP by over $+2.5\%$ for both SWDA~\cite{saito2019strong} and CRDA~\cite{xu2020exploring}. Note that we did not apply FAA to other listed UDA methods due to space limit. 

Table~\ref{table:det_city2BDD} shows domain adaptive object detection over the task Cityscapes $\rightarrow$ BDD100k. It can be seen that including FAA outperforms state-of-the-art UDA methods consistently as in Table~\ref{table:det_city2fog}. Note that we did not apply FAA to other listed UDA methods due to space limit.

\subsection{Domain Adaptive Image Classification}

We presents experimental results on VisDA17 in Table~\ref{table:visda17} in per-class accuracy. It can be seen that incorporating FAA outperforms state-of-the-art UDA methods consistently. This applies to UDA methods that employ stronger backbones ResNet-152~\cite{pinheiro2018unsupervised,sankaranarayanan2018generate}.

Table~\ref{table:office} shows domain adaptive image classification experiments over Office-31 (all using the same backbone ResNet-50). We can see that incorporating FAA leads to robust domain adaptation and improve the image classification consistently by large margins.

\subsection{Discussion}

\begin{figure*}[t]
\begin{tabular}{p{3cm}<{\centering} p{3cm}<{\centering} p{3cm}<{\centering} p{3cm}<{\centering} p{3cm}<{\centering}}
\raisebox{-0.5\height}{\includegraphics[width=1.1\linewidth,height=0.55\linewidth]{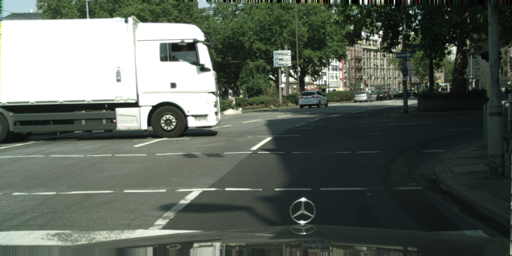}}
& \raisebox{-0.5\height}{\includegraphics[width=1.1\linewidth,height=0.55\linewidth]{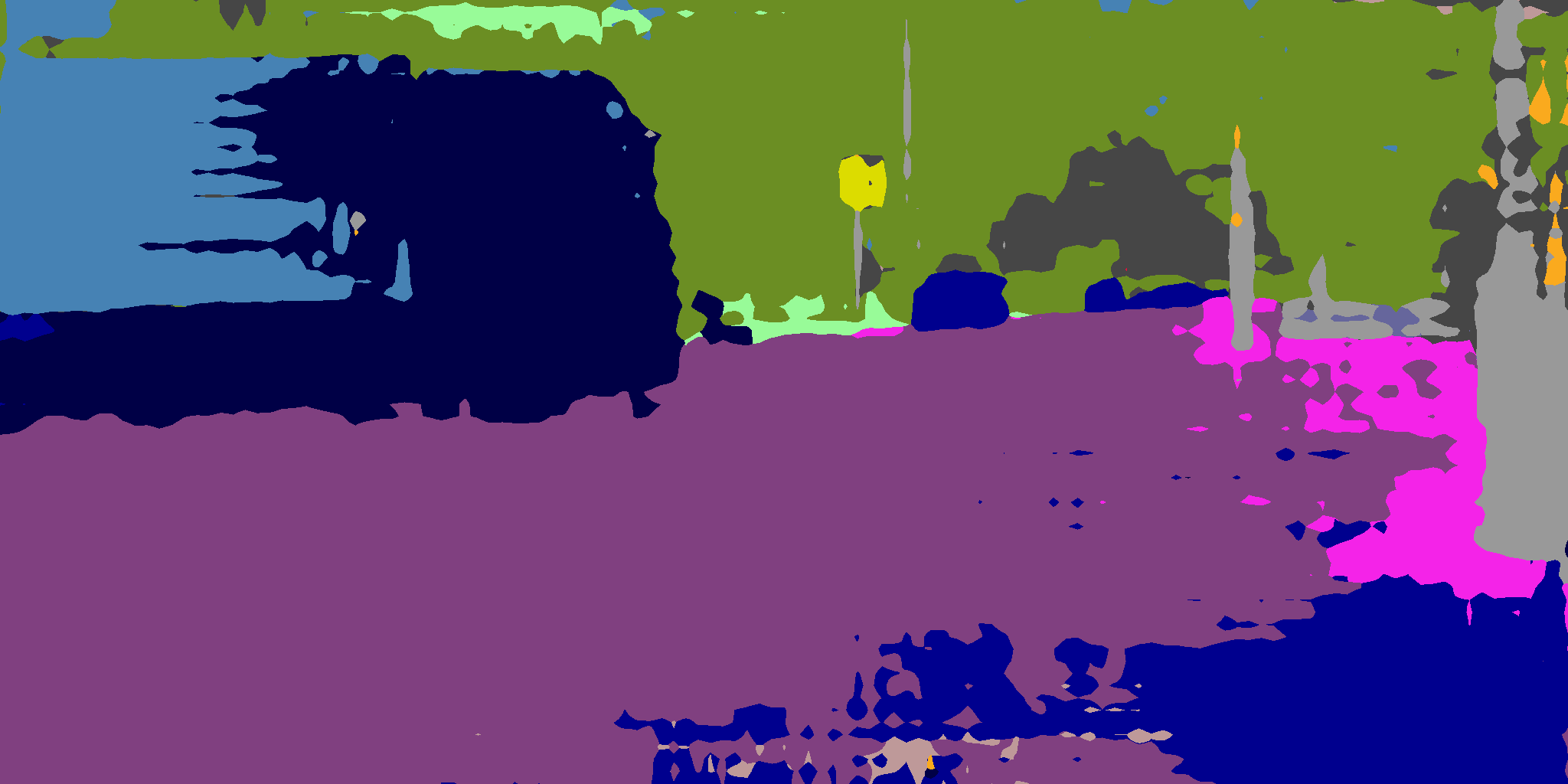}}
& \raisebox{-0.5\height}{\includegraphics[width=1.1\linewidth,height=0.55\linewidth]{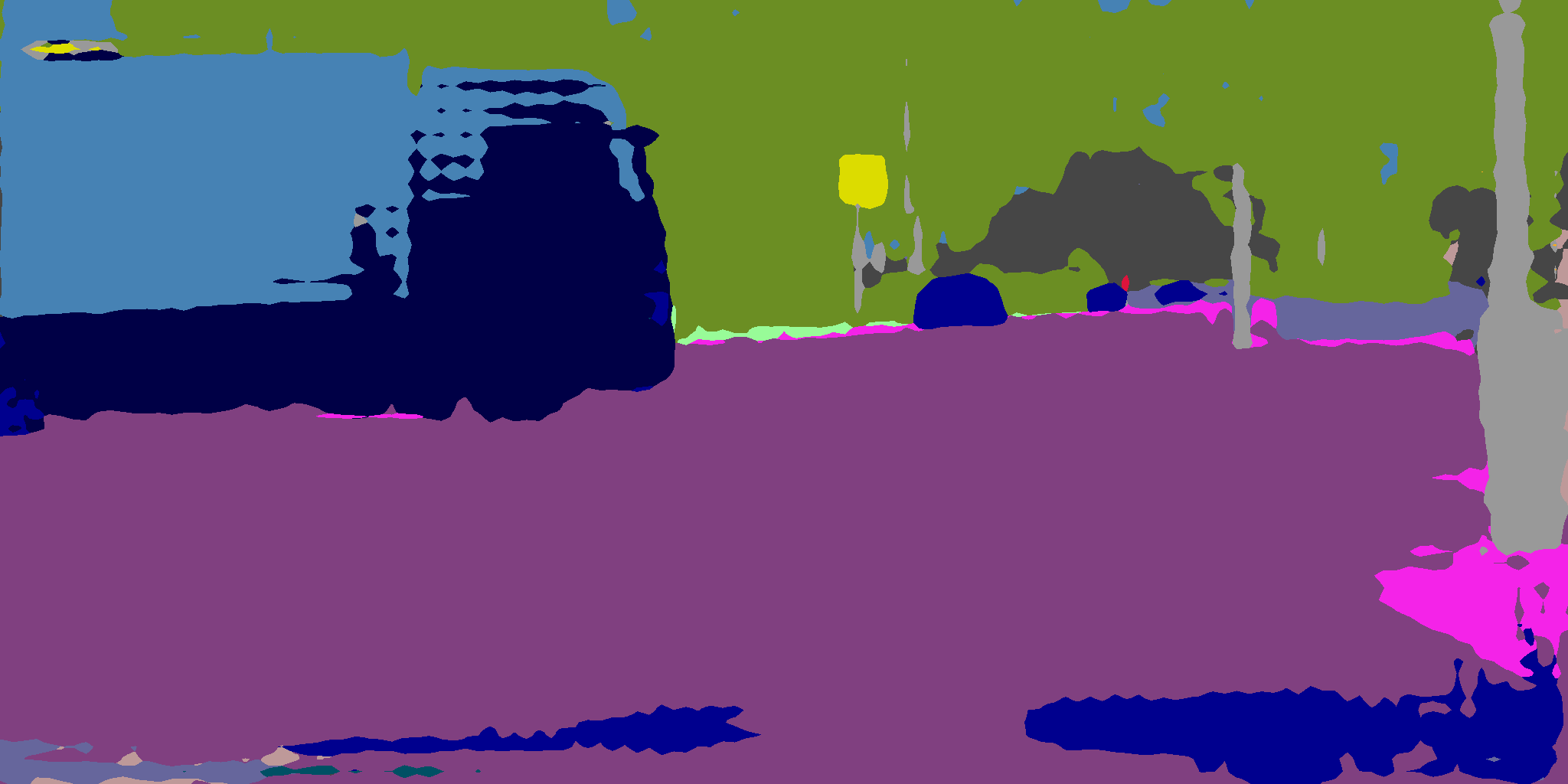}}
& \raisebox{-0.5\height}{\includegraphics[width=1.1\linewidth,height=0.55\linewidth]{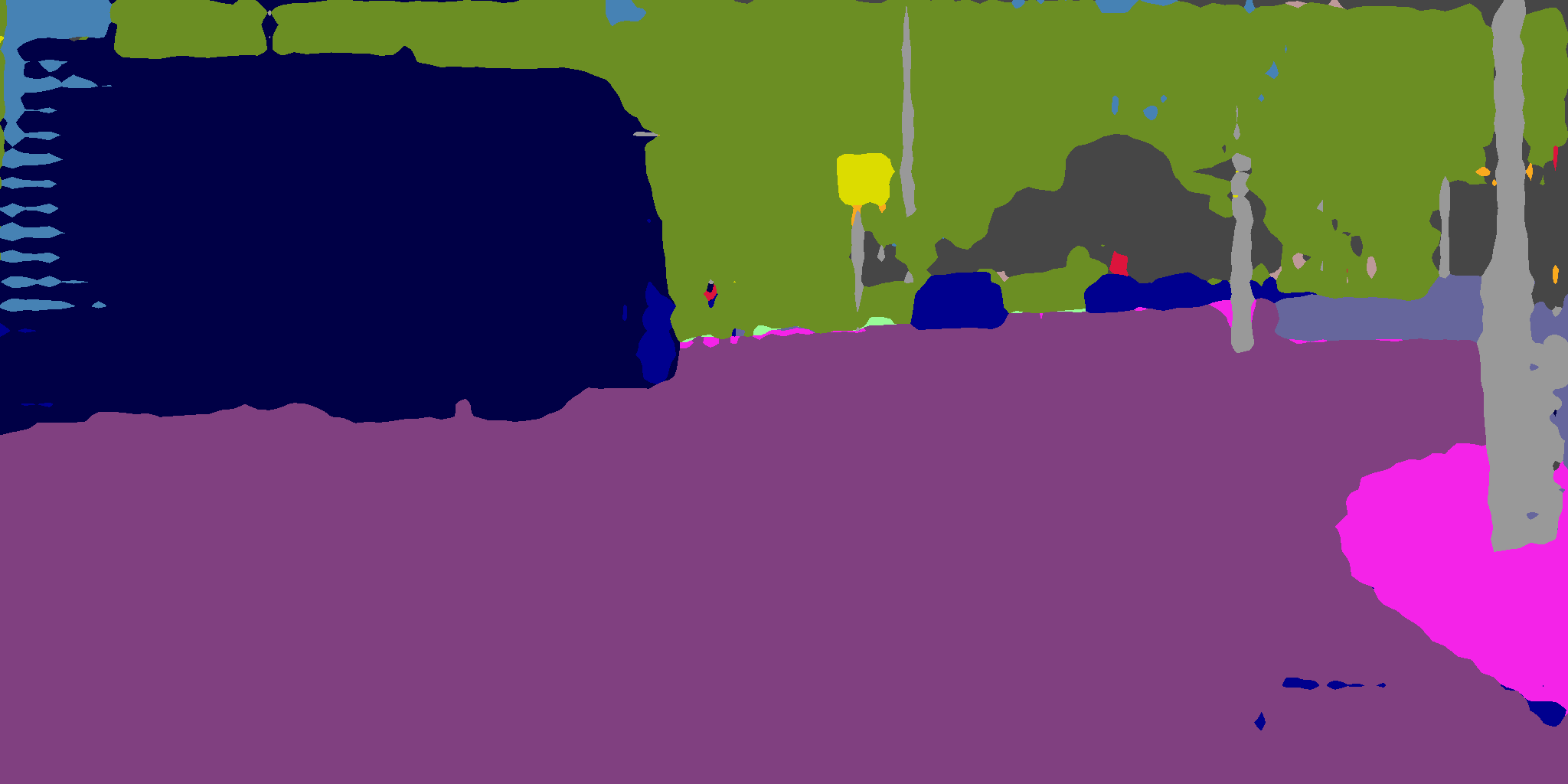}}
& \raisebox{-0.5\height}{\includegraphics[width=1.1\linewidth,height=0.55\linewidth]{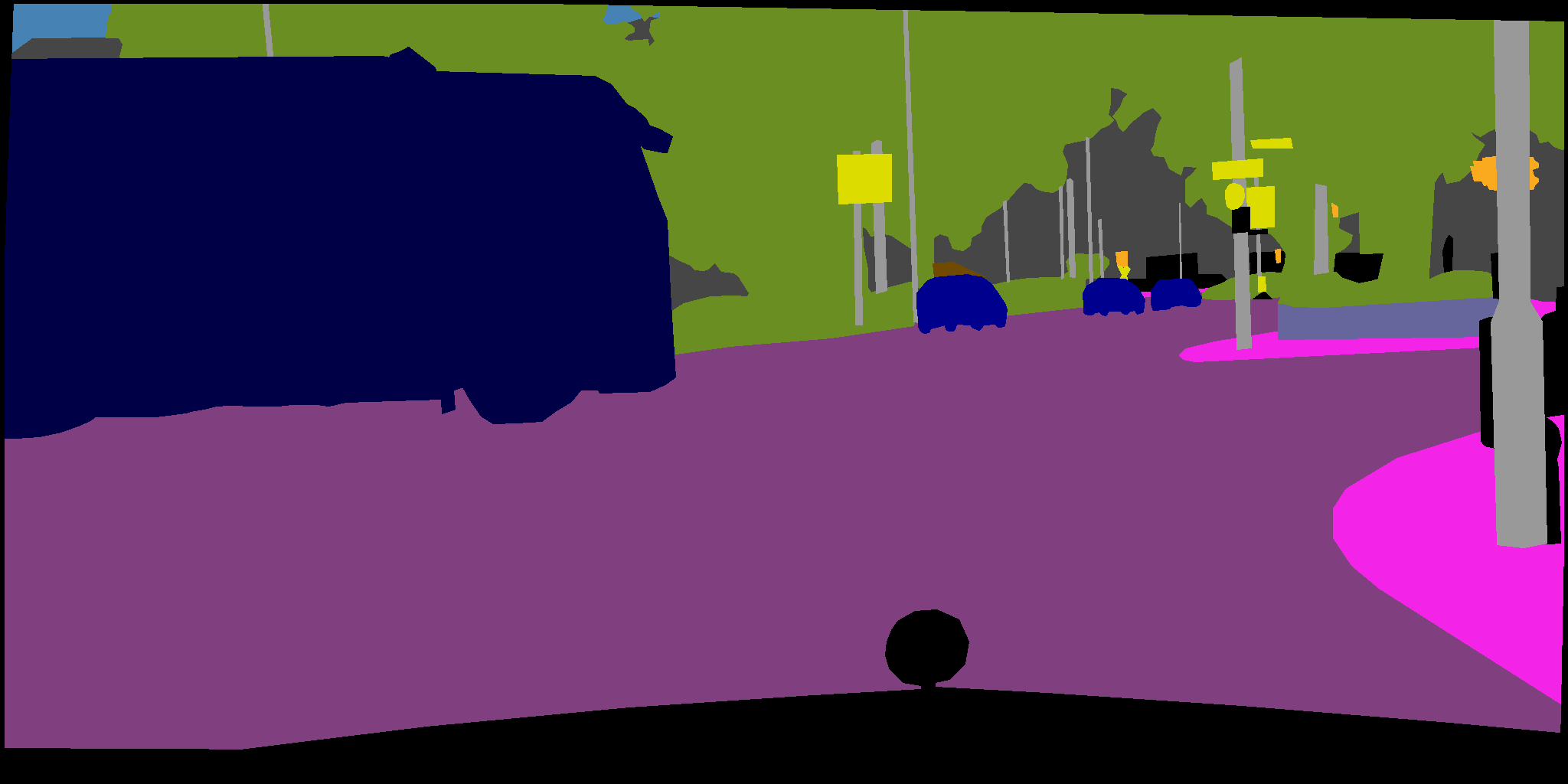}}
\vspace{-2.5 pt}
\\
\raisebox{-0.5\height}{Target Image}
 & \raisebox{-0.5\height}{Baseline}
& \raisebox{-0.5\height}{AdaptSeg~\cite{tsai2018learning}}
& \raisebox{-0.5\height}{\textbf{Ours(AdaptSeg+FAA)}}
& \raisebox{-0.5\height}{Ground Truth}
\\
\end{tabular}
\vspace{2.5 pt}
\caption{
Qualitative illustration of domain adaptive semantic segmentation task GTA5 to Cityscapes task. Our robust domain adaption (RDA) employs Fourier adversarial attacking to mitigate the overfitting in domain adaptive learning effectively, which produces better semantic segmentation by mitigating over-fitting in domain adaptive learning. The differences are clearer for less-frequent categories such as traffic-sign, pole and bus, etc.
}
\label{fig:results_seg_al}
\end{figure*}

\textbf{Qualitative illustration of domain adaptation in segmentation}

We compare our robust domain adaptation (RDA) with the recent domain adaptation method~\cite{tsai2018learning} qualitatively over semantic segmentation task. As Fig.~\ref{fig:results_seg_al} shows, the proposed RDA outperforms the baselines clearly.

\textbf{Overfitting Mitigation:} We compared FAA with existing overfitting mitigation methods. Most existing methods address overfitting through certain network regularization by noise injection to hidden units (\textit{Dropout}), label-dropout (\textit{Label smooth}), gradient ascent (\textit{Flooding}), data and label mixing (\textit{Mixup}), gradient based adversarial attacking (\textit{FGSM}), virtual-label based adversarial attacking (\textit{VAT}), etc. Table~\ref{tab:comp_overfitting} shows experimental results over the task GTA$\rightarrow$Cityscapes. It can be seen that existing regularization does not perform well in the domain adaptation task. The major reason is that existing methods were designed for supervised and semi-supervised learning where training and test data usually have little domain gap. The proposed FAA mitigates overfitting with clear performance gains as it allows large magnitude of perturbation noises
which is critical to the effectiveness of its generated adversarial samples due to the existence of `domain gaps’ in UDA.

Due to the space limit, we provide more visualization of the qualitative segmentation examples (including their comparisons with the recent domain adaptation method~\cite{tsai2018learning}) in the supplementary material.

\begin{table}[t]
\centering
\begin{footnotesize}
\begin{tabular}{c|ccc}
\hline
Method &mIoU  & gain
\\\hline
ST~\cite{zou2018unsupervised} &42.6 &N.A.
\\\hline
+Dropout~\cite{srivastava2014dropout} &43.1 &+0.5\\
+Label smooth~\cite{szegedy2016rethinking} &43.4 &+0.8\\
+Mixup~\cite{zhang2017mixup} &43.6 &+1.0\\
+FGSM~\cite{goodfellow2014explaining} &43.9 &+1.3\\
+VAT~\cite{miyato2018virtual} &44.3 &+1.7\\
+Flooding~\cite{ishida2020we} &44.1 &+1.5\\
\textbf{+FAA} &\textbf{50.1} &\textbf{+7.5}\\
\hline
\end{tabular}
\end{footnotesize}
\caption{
Comparison with existing overfitting migitation methods: For the semantic segmentation task GTA $\rightarrow$ Cityscapes, FAA performs the best consistently by large margins.
}
\label{tab:comp_overfitting}
\end{table}

\section{Conclusion}

In this work, we presented RDA, a robust domain adaptation technique that mitigates overfitting in UDA via a novel Fourier adversarial attacking (FAA).
We achieve robust domain adaptation by a novel Fourier adversarial attacking (FAA) method that allows large magnitude of perturbation noises but has minimal modification of image semantics.
With FAA-generated adversarial samples, the training can continue the `random walk’ and drift into an area with a flat loss landscape, leading to more robust domain adaptation.
Extensive experiments over multiple domain adaptation tasks show that RDA can work with different computer vision tasks (\ie, segmentation, detection and classification) with superior performance. We will explore disentanglement-based adversarial attacking and its applications to other computer vision tasks. We will also study how FAA could mitigate over-fitting in classical supervised learning and semi-supervised learning as well as the recent contrast-based unsupervised representation learning.

\section*{Acknowledgement}
This study is supported under the RIE2020 Industry Alignment Fund – Industry Collaboration Projects (IAF-ICP) Funding Initiative, as well as cash and in-kind contribution from Singapore Telecommunications Limited (Singtel), through Singtel Cognitive and Artificial Intelligence Lab for Enterprises (SCALE@NTU).

{\small
\bibliographystyle{ieee_fullname}
\bibliography{egbib}
}

\end{document}